\documentclass[10pt,journal]{IEEEtranTCOM}

\usepackage[english]{babel}
\usepackage[usenames]{color}
\usepackage[cp1250]{inputenc}
\usepackage{amsfonts}
\usepackage{amsthm}
\usepackage{graphicx}
\usepackage{epsfig}
\usepackage{mathrsfs}
\usepackage{amsmath}
\usepackage{algorithm}
\usepackage{algorithmic}

\theoremstyle{plain}

\begin{document}
\newcommand{\bea}{\begin{eqnarray}}
\newcommand{\eea}{\end{eqnarray}}
\newcommand{\be}{\begin{equation}}
\newcommand{\ee}{\end{equation}}
\newcommand{\beas}{\begin{eqnarray*}}
\newcommand{\eeas}{\end{eqnarray*}}
\newcommand{\bs}{\backslash}
\newcommand{\bc}{\begin{center}}
\newcommand{\ec}{\end{center}}
\def\SC {\mathscr{C}}

\title{Gaussian AutoEncoder}
\author{\IEEEauthorblockN{Jarek Duda}\\
\IEEEauthorblockA{Jagiellonian University,
Golebia 24, 31-007 Krakow, Poland,
Email: \emph{dudajar@gmail.com}}}
\maketitle

\begin{abstract}
Generative AutoEncoders require a chosen probability distribution in latent space, usually multivariate Gaussian. The original Variational AutoEncoder (VAE) uses randomness in encoder - causing problematic distortion, and overlaps in latent space for distinct inputs. It turned out unnecessary: we can instead use deterministic encoder with additional regularizer to ensure that sample distribution in latent space is close to the required. The original approach (WAE) uses Wasserstein metric, what required comparing with random sample and using an arbitrarily chosen kernel. Later CWAE finally derived a non-random analytic formula by averaging $L_2$ distance of Gaussian-smoothened sample over all 1D projections. However, these arbitrarily chosen regularizers do not lead to Gaussian distribution. 

This article proposes approach for regularizers directly optimizing agreement between empirical distribution function and its desired CDF for chosen properties, for example radii and distances for Gaussian distribution, or coordinate-wise, to directly attract this distribution in latent space of AutoEncoder. We can also attract different distributions with this general approach, for example latent space uniform distribution on $[0,1]^D$ hypercube or torus would allow for data compression without entropy coding, increased density near codewords would optimize for the required quantization.
\end{abstract}
\section{Introduction}
Generative AutoEncoders require probability distribution in the latent space being close to a chosen (prior) distribution, usually multivariate Gaussian $N(\mathbf{0},\mathbf{I})$ in $D$-dimensional latent space. The original Variational AutoEncoders (VAE)~\cite{vae} use nondeterministic encoder - choosing from a Gaussian distribution for each input, optimized to minimize Kullback-Leibler distance/divergence for separate inputs. Such randomness means additional distortion, these Gaussians overlap - distinct inputs can lead to the same outputs. Separate treatment lacks tendency for uniform coverage by the sample, which requires some repulsion.

These problems were repaired later by philosophy introduced in WAE article~\cite{wae}. As in standard AutoEncoder, it uses deterministic encoder $(\mathcal{E}:X\to Z)$ minimizing reconstruction cost: distortion of encoding-decoding$(\mathcal{D})$ process - some average over $i$ of distance between $x_i$ and $\mathcal{D}(\mathcal{E}(x_i))$, preferably alongside evaluation of a trained discriminator (GAN) - exploiting the fact that not all distortions are equally unwanted. Additionally, the minimized criterion contains also regularizer - some distance between distribution of  $\{z_i\}_i=\{\mathcal{E}(x_i)\}_i$ obtained ensemble in the latent space and the Gaussian distribution we would like to reach. Assume the number of considered points is $n$, which can be the entire sample size, or size of a used random subset.

\begin{figure}[t!]
    \centering
        \includegraphics{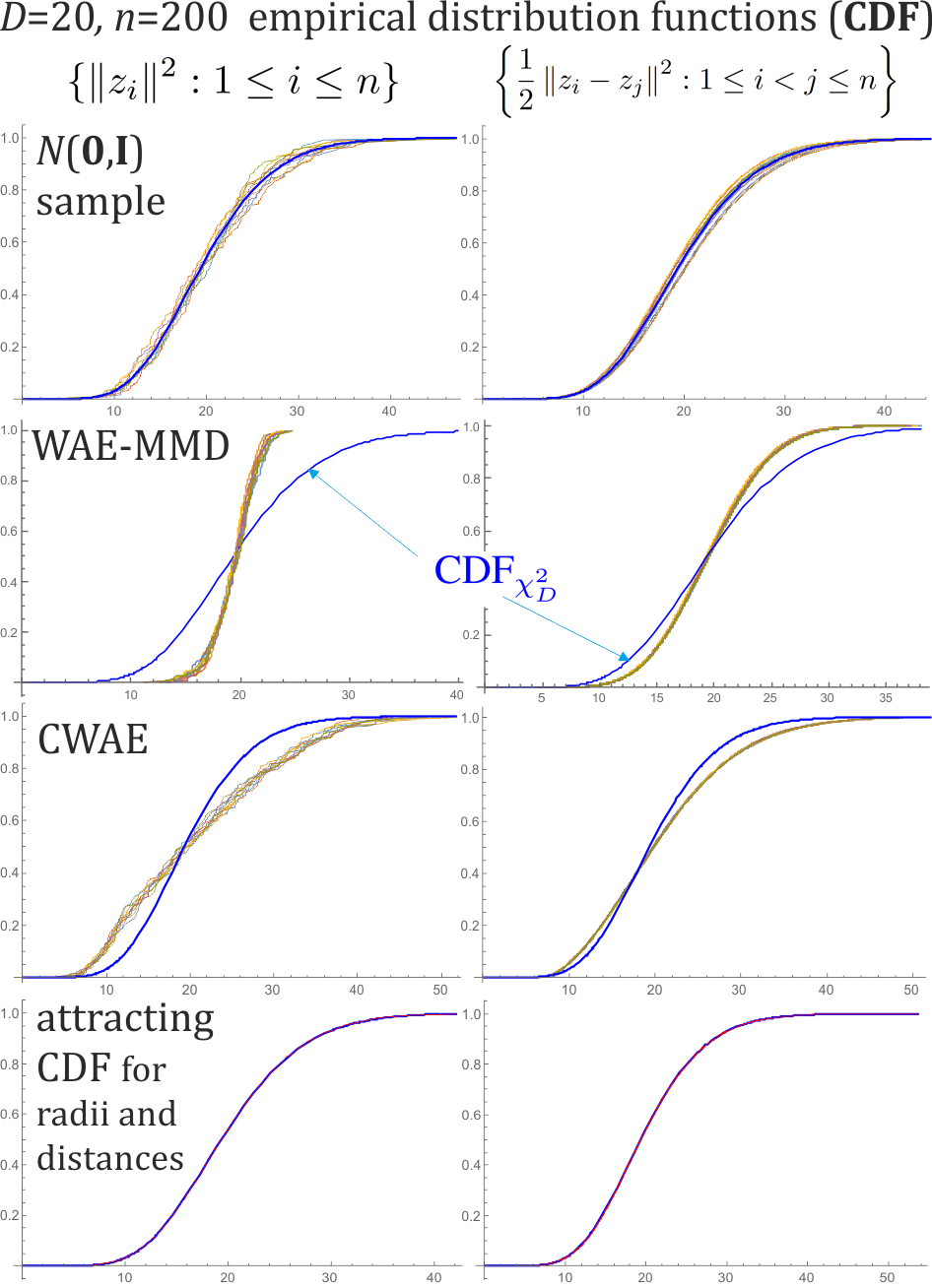}
        \caption{Empirical distribution function (estimated CDF) from sorted squared radii (left column) and distances (right column) for $n=200$ points in $\mathbb{R}^D$ for $D=20$. For independent variables from multivariate Gaussian distribution both should be close to CDF of $\chi^2_D$ distribution. Top row: plots for 10 independent experiments using random sample from $N(\mathbf{0},\mathbf{I})$. 2nd and 3rd row: plots of 10 independent experiments for gradient descent minimization (starting with random sample from uniform distribution in $[-1,1]^D$) of regularizer of WAE-MMD~(\ref{wae}) or CWAE~(\ref{cwae}) formula - obtained distribution is essentially narrower or wider than for Gaussian. Bottom row: discussed here attracting to the desired CDF for radii and distances - getting nearly perfect agreement (hence also of their densities), also in further tests presented in Fig. \ref{tests}. Such optimization step in generative AutoEncoder should be combined with optimization of encoding-decoding distortion and discriminator of decoded vectors.}
       \label{cdffig}
\end{figure}
Such two complex criteria (reconstruction cost and regularizer) are usually evaluated while optimized combined, however, proper evaluation should start with separating them - due to complexity, dependence on data sample, and freedom of choice e.g. of regularization rate. Hence, we will focus here on finding a proper regularizer: which optimization indeed approaches the desired e.g. Gaussian distribution, what turns out quite difficult as we can see in Fig. \ref{cdffig} - often is not satisfied due to focusing on some arbitrary criteria instead of what is really required. This article repairs it: designing regularizers directly attracting toward the desired distribution, to be combined with optimization of reconstruction cost.

The original WAE chooses to optimize approximation of Wasserstein metric, also known as earth mover's distance. The minimized regularizer in WAE-MMD is:
\be \frac{1}{n(n-1)}\sum_{i\neq j} k(z_i, z_j)-\frac{2}{n^2}\sum_{i,j} k(z_i,\tilde{z}_j)\label{wae}\ee
where $\{\tilde{z}_i\}_{i=1..n}$ is a sample for $N(\mathbf{0},\mathbf{I})$ - chosen randomly in every optimization step. Regarding choice of used kernel, the article briefly mentions $k(x,y)=\exp(-\|x-y\|^2)$, then arbitrarily choose to use $k(x,y)=2D/(2D+\|x-y\|^2)$ kernel instead. In the next section we will analytically derive similar formula as for the former choice, Fig. \ref{cdffig} show results of minimization using the latter choice as in the article.

Later sliced SWAE~\cite{swae} uses a different approximation of Wasserstein metric - for randomly chosen 1D projections with again randomly chosen sample and arbitrarily chosen transportation cost.

Soon after it, finally a non-random analytical formula was proposed as CWAE~\cite{cwae} by using $L_2$ distance for KDE (kernel density estimation) Gaussian-smoothened 1D projections and averaging over all projection directions. Its regularizer does not longer require a random sample, getting similar formula as (\ref{wae}) but with reduced one index:
\be\frac{1}{n^2} \sum_{i,j}\frac{1}{\sqrt{\gamma_n +\frac{\|z_i-z_j\|^2}{2D-3}}}-\frac{2}{n}
\sum_i\frac{1}{\sqrt{\gamma_n+\frac{1}{2}+\frac{\|z_i\|^2}{2D-3}}}\label{cwae}\ee
for $\gamma_n=\left(\frac{4}{3n}\right)^{2/5}$ heuristic choice. This formula uses approximation claimed in the article to be practically indistinguishable for tested $D=20$ dimensions. Formulas to directly use multivariate Gaussians instead (without projection) are derived in Section \ref{s2} of this article for general covariance matrices.\\

As above regularizers contain arbitrary choices, randomness and approximations, we should verify if minimization of such regularizer alone indeed leads to $N(\mathbf{0},\mathbf{I})$ Gaussian distribution, as combined with optimization of reconstruction cost it will be even more difficult. It was tested using so called Marida tests~\cite{marida} for 3-rd ad 4-th moment: that $ \frac{1}{n^2} \sum_{i,j=1}^n (z^T_i z_j)^3 \approx 0$ and $\frac{1}{n}\sum_{i=1}^n \|z_i\|_2^4\approx D(D+2)$.

However, these are only two moments, still leaving huge freedom for disagreement with the desired continuous distribution, starting with the second moment $\frac{1}{n}\sum_{i=1}^n \|z_i\|_2^2\approx D$, which generally does not need to be satisfied due to additional constraints (encoding-decoding distortion and evaluation by discriminator). Focusing on them, we could directly optimize for agreement of such moments e.g. using gradient descent. However, from one side it would potentially need infinite number of moments for perfect agreement, from the other choosing weights for separate moments seems a difficult problem.

Much more accurate approach can be found in Kolmogorov-Smirnov test: of some distance between the desired CDF (cumulative distribution function) and empirical distribution function. For Gaussian distribution we would mostly expect agreement of two distributions: for radii $\|z_i\|^2$ and pair-wise distances $\|z_i-z_j\|^2/2$. Both should be from $\chi^2_D$ chi-squared distribution, what turned out not true for WAE-MMD and CWAE regularizer as we can see in Fig. \ref{cdffig} - leading to essentially narrower or wider distribution.\\

Having such accurate criterion we can directly optimize it: agreement of CDF with empirical distribution for chosen properties, especially radii and distances for Gaussian distribution. It will be described in Section \ref{s3} and leads to agreement also for tests of other properties, like random projections, scalar products and distances between normalized vectors - presented in Fig. \ref{tests}. Alternative approach is optimizing distribution of simultaneously all coordinates this way.

Therefore, combining or interleaving it with minimization of reconstruction cost of AutoEncoder (instead of optimizing some arbitrary criterion), we can get direct attraction to Gaussian distribution for latent variable. We can analogously use this approach to attract a different chosen distribution by selecting its crucial 1D properties and directly attracting their proper CDFs. For example enforcing uniform distribution on $[0,1]^D$ hypercube or torus would allow for data compression without additional statistical analysis and entropy coding. Additionally, to optimize for unavoidable quantization, we can enforce increased density near codewords this way.

Section \ref{s2} presents approach of the first version of this article, giving some connection between (\ref{wae}) and (\ref{cwae}) formula, which can have also some other applications like optimization of Gaussian mixture models (GMMs). Section \ref{s3} contains the main proposed approach.
\section{$L_2$ Gaussian mixture distance} \label{s2}
In this section there are derived analytic formulas for $L_2$ distance between multivariate Gaussian-smoothened samples, also using general covariance matrices. The derived formulas are similar to (\ref{wae}) and (\ref{cwae}), can be useful in low dimensions, also e.g. to optimize GMMs. However, high dimensional Gaussians should be rather imagined as thin shells instead of balls - what will be resolved in the next section by directly ensuring agreement of CDF for radii and distances.
\subsection{Integral of product of multivariate Gaussians: $d_{\mu,\Sigma,\Gamma}$}
Density of multivariate $D$-dimensional Gaussian distribution $N(\mu,\Sigma)$: with $\mu$ center and $\Sigma$ covariance matrix (real, symmetric, positive-definite, e.g. $|\Sigma|\equiv\det \Sigma>0$) is:
\be\rho_{\mu, \Sigma}(x):= \frac{1}{\sqrt{|2\pi\Sigma|}}
e^{-\frac 12 (x-\mu)^T\Sigma^{-1} (x-\mu)} \ee
We first need to calculate formula for integral of product of two such densities: of covariance matrix $\Sigma$ and $\Gamma$, which are shifted by a vector $\mu$. Due to translational invariance, we can choose centers of these Gaussians as $\mu$ and $\mathbf{0}:=(0,\ldots,0)$:

\be d_{\mu,\Sigma,\Gamma}:=\int_{\mathbb{R}^D} \rho_{\mu, \Sigma}(x)\cdot\rho_{\mathbf{0},\Gamma}(x)\,dx = \label{form1}\ee
$$\frac{1}{(2\pi)^D\sqrt{|\Sigma||\Gamma|}}\int_{\mathbb{R}^D}
e^{-\frac{(x-\mu)^T \Sigma^{-1}(x-\mu)+x^T\Gamma^{-1}x}{2}}dx$$
Transforming the numerator in exponent we get:
$$\alpha:=x^T(\Sigma^{-1}+\Gamma^{-1})x-2x^T \Sigma^{-1} \mu +\mu^T \Sigma^{-1}\mu$$
Denoting $\Pi=(\Sigma^{-1}+\Gamma^{-1})^{-1}\ $ and $\ \nu=\Pi \Sigma^{-1} \mu\ $ we get:
$$\alpha=x^T \Pi^{-1} x - 2x^T \Pi^{-1} \nu +\mu^T \Sigma^{-1} \mu =$$
$$=(x-\nu)^T \Pi^{-1} (x-\nu) -\nu^T \Pi^{-1} \nu +\mu^T \Sigma^{-1} \mu $$
Now knowing that
$\frac{1}{\sqrt{|2\pi\Pi|}}\int_{\mathbb{R}^D} e^{-\frac 12 (x-\nu)^T\Pi^{-1} (x-\nu)}\, dx = 1 $
we can remove integral from (\ref{form1}):
$$d_{\mu,\Sigma,\Gamma} = \frac{\sqrt{(2\pi)^D|\Pi|}}{(2\pi)^D\sqrt{|\Sigma||\Gamma|}}
e^{\frac 12 \left(\nu^T \Pi^{-1} \nu-\mu^T\Sigma^{-1} \mu \right)}$$
Substituting $\Pi=(\Sigma^{-1}+\Gamma^{-1})^{-1},\ \nu=\Pi \Sigma^{-1} \mu$ :
$$\nu^T \Pi^{-1} \nu-\mu^T\Sigma^{-1} \mu=\mu^T\Sigma^{-1}\Pi\,\Pi^{-1}\Pi\,\Sigma^{-1}\mu-\mu^T\Sigma^{-1} \mu=$$
$$\mu^T\Sigma^{-1}\left(\Pi\Sigma^{-1}-\Pi\, \Pi^{-1}\right)\mu=
-\mu^T\Sigma^{-1}(\Sigma^{-1}+\Gamma^{-1})^{-1} \Gamma^{-1}\mu$$
Observe that, as required, it does not change if switching $\Sigma$ and $\Gamma$. We can now get the final formula:

\be d_{\mu,\Sigma,\Gamma}=
\frac{\exp\left(-\frac 12 (\mu^T\Sigma^{-1}(\Sigma^{-1}+\Gamma^{-1})^{-1} \Gamma^{-1}\mu) \right)}
{\sqrt{(2\pi)^D |\Sigma||\Gamma||\Sigma^{-1}+\Gamma^{-1}|}}\ee
Let us also find its special case for spherically symmetric Gaussians: $\overline{d}_{l,\sigma^2,\gamma^2}:= d_{l\hat{\mu},\sigma^2 \mathbf{I},\gamma^2 \mathbf{I}}$ for any length 1 vector $\hat{\mu}$:
$$\overline{d}_{l,\sigma^2,\gamma^2}=
\frac{\exp\left(-\frac 12 \, l^2 \, \sigma^{-2}(\sigma^{-2}+\gamma^{-2})^{-1}\gamma^{-2}\right)}
{\sqrt{(2\pi)^D \left(\sigma^{2}\gamma^{2}(\sigma^{-2}+\gamma^{-2})\right)^D}}$$
\be \overline{d}_{l,\sigma^2,\gamma^2}= d_{l\hat{\mu},\sigma^2 \mathbf{I},\gamma^2 \mathbf{I}}=\frac{\exp \left(-\frac 12 \frac {l^2}{\sigma^2+\gamma^2} \right)}
{\sqrt{2\pi \left(\sigma^{2}+\gamma^{2}\right)}^D} \ee

We could also analogously find formula for integral of three or more Gaussians. We can also use general powers of Gaussians, e.g. to calculate $L_p$ norm, for example using: $(\rho_{\mu,\Sigma})^p= \sqrt{|2\pi\Sigma|}^{1-p} p^{-D/2}\, \rho_{\mu,\Sigma/p}$.
\subsection{$L_2$ distance between two smoothened samples}
Having two samples  $X=(x_i)_{i=1..n}$ and $Y=(y_j)_{j=1..m}$ in $\mathbb{R}^D$, we would like to KDE smoothen them using multivariate Gaussians, then define distance as $L_2$ norm between such smoothened samples.

For full generality, let us start with assuming that each point has a separately chosen covariance matrix for the Gaussian: we have some $(\Sigma_i)_{i=1..n}$ and $(\Gamma_j)_{j=1..m}$ matrices. Such Gaussian mixture can use any positive weights $(w_i)_{i=1..n},\ (v_i)_{j=1..m}$ summing to 1, for simplicity we can assume that they are equal $w_i=1/n$, $v_j=1/m$.

Now such squared $L_2$ distance between these samples, depending on the choice of covariance matrices, is
$$ d^2_g(X,Y)=
\int_{\mathbb{R}^D} \left(\sum_i w_i\rho_{x_i,\Sigma_i}-\sum_j v_j \rho_{y_j,\Gamma_j}\right)^2 dx= $$
$$\sum_{i,i'=1}^n \frac{d_{x_i-x_{i'},\Sigma_i,\Sigma_{i'}}}{n^2}+
\sum_{j,j'=1}^m \frac{d_{y_i-y_{i'},\Gamma_j,\Gamma_{j'}}}{m^2}-
2\sum_{i=1}^n \sum_{j=1}^m \frac{d_{x_i-y_j,\Sigma_i,\Gamma_j}}{nm}
$$
As we have freedom of choosing $(\Sigma_i)_{i=1..n}, (\Gamma_i)_{i=1..m}$, we can use above formula to optimize this choice - we can use distance being a result of e.g. its iterative minimization. The initial choice can be found with mean-field approximation discussed later.

This formula can  be used for example for optimizing GMM (Gaussian Mixture Model) - e.g. associate fixed Gaussians to points of the sample and find $L_2$ close covering with a smaller number of Gaussians. It allows to directly optimize centers and covariances matrices: as symmetric $\Sigma^{-1}$ (so called precision matrix) for efficient calculation, or as $\Sigma=O^T D O$.

Let us also find more practical formula for the basic choice: of all covariance matrices being $\sigma^2 \mathbf{I}$:
\be \overline{d}^2_g(X,Y)\cdot \sqrt{4\pi \sigma^{2}}^D=\ee
$$\sum_{i,i'=1}^n
\frac{e^{- \frac{\|x_i-x_{i'}\|^2}{4\sigma^2}}}{n^2}+\sum_{j,j'=1}^m \frac{e^ {-\frac {\|y_j-y_{j'}\|^2}{4\sigma^2}}}{m^2}-2\sum_{i=1}^n \sum_{j=1}^m \frac{e^{-\frac {\|x_i-y_{j}\|^2}{4\sigma^2}}}{nm}$$

We can remove the $\sqrt{4\pi \sigma^{2}}^D$ fixed term while applying this formula - as it becomes very large in high dimensions.

This formula turns out quite similar as for WAE (\ref{wae}) with exponential kernel, using second sample as random from the chosen distribution. In the next subsection we will directly use a single Gaussian instead - getting similar formula as final for CWAE (\ref{cwae}). It uses another, heavy tailed kernel: $\phi_D(s)\approx \left(1+4s/(2D-3)\right)^{-1/2}$ function in place of exponent. Similarity with CWAE comes from similar origin: both use $L_2$ distance between Gaussian-smoothened samples. However, CWAE calculates this distance for projections to 1D subspaces and averaging over all such directions - optimizing similarity of 1D projections. In contrast, here we directly want closeness of multivariate distributions - as in the original generative AutoEncoder motivation.

It might be also worth to explore different types of tails - corresponding to repulsion inside both sets, and attraction between them. Like they were charged with various types of Coulomb-like interaction.
\subsection{$L_2$ distance between smoothened sample and $N(\mathbf{0},\mathbf{I})$}
For generative AutoEncoders we are more interested in calculating distance from single Gaussian distribution $N(\mathbf{0},\mathbf{I})$, instead of representing it with a random sample like in WAE. Let us now use $N(\mathbf{0},\mathbf{I})$ in place of $Y$ from the previous subsection:
$$ d^2_g(X,N(\mathbf{0},\mathbf{I}))=
\int_{\mathbb{R}^D} \left(\frac{1}{n}\sum_i \rho_{x_i,\Sigma_i}(x)-\rho_{\mathbf{0},\mathbf{I}}(x)\right)^2 dx= $$
\be\frac{1}{n^2} \sum_{i,i'=1}^n d_{x_i-x_{i'},\Sigma_i,\Sigma_{i'}}+
\frac{1}{\sqrt{4\pi}^D}-
\frac{2}{n}\sum_{i=1}^n d_{x_i,\Sigma_i,\mathbf{I}}
\label{cov} \ee

Using the simplest: spherically symmetric $\Sigma_i=\sigma_i^2\mathbf{I}$, for example for constant $\sigma_i=\sigma$, we get:
$$ \overline{d}^2_g(X,N(\mathbf{0},\mathbf{I}))\cdot \sqrt{4\pi}^D=$$
\be\sum_{i,i'=1}^n
\frac{e^{- \frac{\|x_i-x_{i'}\|^2}{2(\sigma_i^2+\sigma_{i'}^2)}}}{n^2\sqrt{(\sigma_i^2+\sigma_{i'}^2)/2}^D}+1-
\frac{2}{n}\sum_{i=1}^n \frac{e^{- \frac{\|x_i\|^2}{2(1+\sigma_i^2)}}}{\sqrt{(1+\sigma_i^2)/2}^D}\label{sym}\ee
For large $D$ it requires to use $\sigma=1+\epsilon$, for tiny $\epsilon\geq 0$ allowing to manipulate weight of the two above sums. For the simplest choice: $\sigma=1$, formula (\ref{sym}) becomes inexpensive:
\be 1+\frac{1}{n}+\frac{2}{n^2}\sum_{i<i'}
e^{- \frac{\|x_i-x_{i'}\|^2}{4}}-\frac{2}{n}\sum_{i=1}^n e^{- \frac{\|x_i\|^2}{4}}\label{sig1}\ee



\subsection{Mean-field approximation for optimizing $\sigma(\|x\|)$}
Choosing $\sigma$ is generally a difficult question, but we can use kind of mean-field approximation to individually choose covariance matrices depending on position. Specifically, focusing on a given point $x\in X$, we can assume that the remaining ones are from approximately the desired $N(\mathbf{0},\mathbf{I})$ density. This way e.g. $n\,\overline{d}^2_g(X,N(\mathbf{0},\mathbf{I}))$ distance becomes:
$$n\int_{\mathbb{R}^D} \left(\left(\frac{1}{n} \rho_{x,\sigma}(y) +\frac{n-1}{n} \rho_{\mathbf{0},\mathbf{I}}(y)\right) -\rho_{\mathbf{0},\mathbf{I}}(y)\right)^2 dy= $$
\be=\frac{1}{\sqrt{4\pi \sigma^{2}}^D}+\frac{1}{\sqrt{4\pi }^D}-2\frac{e^{- \frac{\|x\|^2}{2(1+\sigma^2)}}}{\sqrt{2\pi(1+\sigma^2)}^D}\label{mf}\ee
For fixed $D$, we would like to choose $\sigma(\|x\|)$ minimizing (\ref{mf}) depending on radius $r=\|x\|$. Obviously, $\sigma(0)=1$. Numerically, approximate behavior turns out \be\sigma(r)\approx 1+\frac{r^2}{2D}\ee
which can be used as $\sigma_i=\sigma(\|x_i\|)$ in distance (\ref{sym}).

This mean-field approximation can be also used to choose optimized position-dependent general covariance matrix: $\Sigma(x)$. Due to symmetry, it should have only two different eigenvalues: in $x$ direction, and in its perpendicular plane.
\subsection{High dimensional situation}
Above calculations might be useful in a few dimensional situation, but in practice we often need to work on large $D$.  As $x\sim N(\mathbf{0},\mathbf{I})$ can be seen as $D$ independent variables (coordinates) from $N(0,1)$, hence $\|x\|^2\sim\chi^2_D$ is from chi-squared distribution, which asymptotically (large $D$) is $\approx N(D,2D)$, making exponent e.g. in (\ref{sig1}) impractically small. It got heavier tail in CWAE (\ref{cwae}) by 1D projections (also in WAE (\ref{wae}) but without a deeper explanation).

Hence, high dimensional Gaussian distribution should be rather imagined as thin radius $\sqrt{D}$ spherical shell, what is far from ball-like low dimensional intuition about Gaussian mixtures, above $L_2$ distance should be rather imagined as between spheres - not exactly what we are interested in.
\section{Attracting to a chosen CDF} \label{s3}
The previously discussed approaches tried to guess a metric, hoping it will lead to the Gaussian distribution. Instead, we can focus on features of this distribution and try to directly optimize them. We could use moments for this purpose, but they provide only a very rough description.

\begin{figure}[b!]
    \centering
        \includegraphics{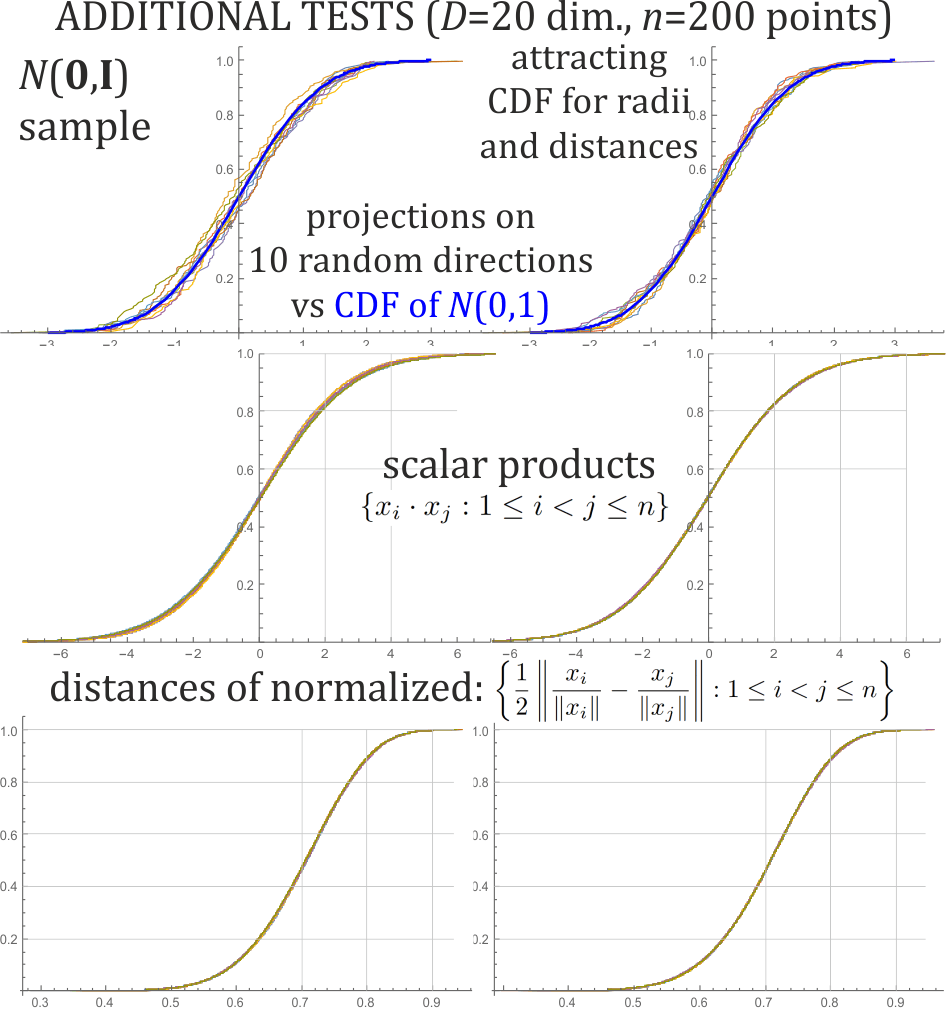}
        \caption{Additional tests for discussed attracting Gaussian CDF for radii and distances (right column) compared with random i.i.d. sample (left column) from distribution we would like to achieve $(N(\textbf{0},\textbf{I}))$ - each contains 10 independent experiments. Top row: test of projections on random directions. Middle row: test of scalar products. Bottom row: test of uniform distribution of angles as distances for normalized vectors.}
       \label{tests}
\end{figure}

In contrast, a perfect description of continuous 1D distribution is given by its CDF, and like in Kolmogorov-Smirnov test, it can be modelled as empirical distribution function - by just sorting the values. The most important 1D properties of multivariate Gaussian, other discussed methods were also focused on, are radii and distances - the provided algorithm directly attracts for their agreement. Analogously there can be added (or chosen from scratch) other properties to optimize. However, it turns out that optimizing radii and distances here also leads to agreement of other properties, as we can see in some tests in Fig. \ref{tests}.
\subsection{Algorithm}
This subsection contains the main approach of this article: directly optimizing agreement with the proper CDF of empirical distribution from the sample - obtained by sorting the values. Its used Mathematica implementation is in Appendix.

The discussed version attracts to CDFs of multivariate Gaussian distribution for two (squared) properties: $n$ radii and their $n':=n(n-1)/2$ pairwise distances, both ideally should be from $\chi^2_D$ chi-squared distribution. This general approach can be naturally modified for agreement of other properties and their chosen CDFs.

\textbf{Algorithm:}

We need first to put into tables the desired CDF arguments, here of chi-squared distribution for radii and distances:
$$c_i = \textrm{CDF}_{\chi^2_D}^{-1}((i-0.5)/n) \qquad \textrm{for}\quad i=1\ldots n $$
$$c'_k = \textrm{CDF}_{\chi^2_D}^{-1}((k-0.5)/n') \qquad \textrm{for}\quad k=1\ldots n' $$

Then gradient descent step for optimizing empirical distribution of $(x_i)_{i=1..n}$ set of points in $\mathbb{R}^D$ is:
\begin{enumerate}
  \item Calculate all $n$ radii and $n'=n(n-1)/2$ distances:
  $$\left(r_i = \|x_i\|^2\right)_{i=1..n}\qquad \left(d_{ij} =\frac{1}{2}\|x_i-x_j\|^2\right)_{1\leq i<j\leq n} $$
  \item Sort both - find orders (bijections):
  $$s:\{1..n\}\rightarrow \{1..n\} \quad :\quad   r_{s(1)}\leq r_{s(2)}\leq \ldots\leq r_{s(n)} $$
  $$ s':\{1..n'\}\to \{ij: 1\leq i<j\leq n\}\quad:$$
  $$ d_{s'(1)}\leq d_{s'(2)}\leq \ldots d_{s'(n')}$$
  \item Assuming the minimized final distance is $\ell_1$, which corresponds to area of difference between the desired CDFs and empirical distributions for radii and distances:
  \be \overline{d}=\frac{1}{n}\sum_{i=1}^n |r_{s(i)}-c_i| + \frac{1}{n'} \sum_{k=1}^{n'} |d_{s'(k)}-c'_k|, \label{distf}\ee
  its gradient on $i$-th vector $x_i$ is:
  \be g_i = \frac{2}{n}\,x_i\,\textrm{sgn}(r_i-c_{s^{-1}(i)}) +  \ee
  $$+\frac{2}{n'} \sum_{k:s'(k)=ij\,\vee\, ji} (x_i - x_j)\cdot \textrm{sgn}(d_{s'(k)}-c'_k) $$
  \item Gradient descent e.g. $\forall_i\  x_i = x_i - \alpha g_i$ where $\alpha$ can be chosen depending (e.g. as proportional) to $\overline{d}$.
\end{enumerate}
Each such 1)-4) iteration takes our points closer to agree with perfect CDF of multivariate Gaussian. In AutoEncoder it should be combined or interleaved with steps reducing distortion of coding-decoding process (preferably also  evaluation of discriminator), regularizer rate $\alpha$ should start large and be gradually reduced during training.

It is tempting to approximate CDF of $\chi^2_D$ with just a step function in $D$ (especially in high dimensions) as it would allow to remove above sorting, just optimize both squared norms to constant value $c=c'=D$. Sorting gives more tolerance for distortion from these constants especially for extreme values, exactly like in the real Gaussian distribution.
\subsection{Some comments and expansions}
Proportion of weights for radii and distances was chosen arbitrarily, what might be worth exploring, especially if adding CDFs of more properties to be attracted.

In Kolmogorov-Smirnov test there is used $\ell_\infty$ norm instead, but optimizing it would lead to gradient descent shifting only single extreme points. Above $\ell_1$ norm allows to optimize all points at a time and has a natural interpretation as area between the two plots. It might be worth exploring also other norms like $\ell_2$, which can be obtained by just replacing above sign with bracket.

Above attraction only ensures approaching the desired CDF for radii and pairwise distances, what turns out sufficient for optimizing regularizer alone, also for some other properties as we can see if Fig. \ref{tests}. Approaching it might turn out more difficult while adding other optimization criteria like reconstruction cost - when it might be worth to consider adding CDF attraction also for other properties like scalar products $x_i\cdot x_j$ or $\|x_i+x_j+x_k\|^2$. If there is a problem with analytical formula for CDF, it can be approximated by just sampling from the desired distribution and using empirical distribution.

This general approach can be also used to attract different chosen distributions in the latent space, what requires choosing essential properties: for which CDF we would like to attract, then replacing above $\overline{d}$ with the chosen sum. For example to attract GMM-like distribution, we can choose agreement of CDF (not necessarily shell-like as in Gaussian) of distances from a few chosen points $\|x_i-\mu_j\|$ like centers in GMM, and CDF for distances $\|x_i-x_j\|$ found e.g. as empirical distribution of random sample.

Alternative approach e.g. for Gaussian is optimizing CDF simultaneously for all coordinates. Assume coordinates have independent distributions, $\textrm{CDF}_j$ is the desired CDF of $j$-th coordinate, e.g. error function for Gaussian, or $\textrm{CDF}_j(v)=v$ for uniform distribution on $[0,1]^D$ hypercube or torus. For $x$ from data sample, denote by $s^j\in\{1\ldots n\}$ its position while sorting accordingly to $j$-th coordinate. Hence \be\tilde{x}=\left(\textrm{CDF}^{-1}_j((s^j-0.5)/n\right)_{j=1\ldots D}\ee is its perfect position accordingly to the desired distribution, we can use e.g. $x\to x+\alpha (\tilde{x}-x)$ as regularization step.
\subsection{Data compression application}
For data compression applications, especially image/video, we would like to learn from dataset how typical objects (e.g. textures) look like and try to encode within their space - which is essentially smaller than the space of e.g. all bitmaps. It is usually realized by encoding crucial features, like Fourier or wavelet coefficients in classical methods. Machine learning techniques can optimize it further - customize based on the training dataset.

Additionally, images usually have patterns repeating in various scales. To exploit this multi-scale nature, there was proposed pyramidal decomposition~\cite{waveone} in analogy to wavelet transform: encode a given block simultaneously in multiple scales: differing by down-sampler. Optimizing distortion of encoding-decoding process (including quantization), and additionally evaluation by discriminator, we get kind of multi-scale AE-GAN, with additional encoding of quantized features - values of latent variables.

In standard AuteEncoder these values of latent variables usually have very complex distribution, making their statistical analysis (and entropy coding) difficult and often suboptimal, what translates into inferior compression ratio. Beside simplification (cost), we should get better compression if enforcing some simple probability distribution in latent space by adding some regularizer to optimized criteria. For example if enforced multivariate Gaussian, each coordinate should be from approximately 1D Gaussian, which can be encoded by splitting possible values into ranges (bins), then use entropy coder to store bin's number, then directly store some number of the remaining most significant bits if needed~\cite{physics}. Alternative approach is using vector quantization, for example separately encode radius, and $\hat{x}=x/\|x\|$ from uniform distribution on unit sphere e.g. using pyramid vector quantizer \cite{pyr1,pyr2}.

We can also use the discussed attracting CDF approach to enforce a different distribution. For example mentioned uniform distribution in $[0,1]^D$ hypercube would allow to avoid entropy coder - we could just directly store a chosen number of the most significant bits for each coordinate. It could be done by analogously attracting to uniform distribution for all coordinates, however, it needs some special behavior (e.g. repulsion, projection $\max(0,\min(1,v))$, or rescaling) near the boundaries (not to exceed them). We could repair it by using $[0,1]^D$ torus instead: gluing pairs of surfaces (in 0 and 1) for all $D$ dimensions, by just taking modulo 1 for each coordinate originally being a real number.

For data compression applications we need to also include quantization ($Q$) of latent space to practically represent and encode these continuous values: as the closest point (codeword) from a chosen finite subset (codebook). Directly including it ($x\to \mathcal{D}(Q(\mathcal{E}(x)))$) during AutoEncoder training would give differential equal zero. A simplest way to resolve this problem is just ignoring quantization during training. More sophisticated solution is adding quantization error: e.g. $\|z-Q(z)\|_2$ to optimized criteria, getting "egg-carton"-like potential with minima in codewords. Discussed here attracting to a chosen CDF allows to include such behavior inside this chosen CDF. For example for uniform distribution on $[0,1]^D$ and taking $k$ most significant bits for each coordinate, instead of $\textrm{CDF}_j(v)=v$, we can e.g. use $\textrm{CDF}_j(v)=2^{-k}\lfloor 2^k v+0.5\rfloor$, getting additional attraction to the closest codeword: point of $s^j\in\{0\ldots n-1\}$ position of $j$-th coordinate should be shifted $x\to x+\alpha (\tilde{x}-x)$ toward $\tilde{x}=\left(2^{-k}\,(\lfloor 2^k s^j/n\rfloor +0.5)\right)_{j=1\ldots D}$.
\section{Conclusions and further work}
The basic conclusion of this article is that instead of using heuristic approximated regularizers, in similar computational cost we can directly optimize toward the desired probability distribution e.g. for radii and distances of multivariate Gaussian distribution. Combining or interleaving such optimization step with standard AutoEncoder optimization (of encoding-decoding distortion and evaluation by discriminator), we can ensure that the final distribution of latent variable is nearly indistinguishable from a random sample from the desired probability distribution.

Beside testing the proposed approach with AutoEncoders, suggested further work starts with expanding evaluation of other methods from just testing of two moments, to much more accurate: verifying agreement of empirical distributions with desired CDFs like in fig. \ref{cdffig}.

As discussed, above approach leaves some freedom which might be worth exploring, e.g. weights between CDFs for different properties, set of these properties, norm for evaluating distance between CDF and empirical distribution.

Finally, this attracting CDF approach is much more general: can be used to approach practically any chosen probability distribution, what allows to use e.g. a chosen clustering in latent space with GMM-like prior distribution, or even distribution with some chosen nontrivial topology e.g. for some circular morphing, or $[0,1]^D$ torus latent space to simplify and optimize storage of its value in data compression, also with "egg-carton"-like density to optimize for quantization.
\appendix
Used Mathematica implementation of approaching chosen radii and distances CDFs:

\begin{footnotesize}
\begin{verbatim}
np = n*(n - 1)/2;                (* n points in R^d *)
(* calculating CDF tables and auxiliary tables: *)
invcdf[q_] := InverseCDF[ChiSquareDistribution[d], q];
c = Table[invcdf[(i - 0.5)/n], {i, n}];
cp = Table[invcdf[(k - 0.5)/np], {k, np}];
dt = Table[0., {i, np}];               (* distances *)
ps = Table[0, {i, n}];        (* positions in order *)
psp = Table[0, {i, np}];

(* single optimization step for x table *)
g = Table[0., {i, n}, {j, d}];    (* gradient table *)
rt = Table[Total[x[[i]]^2], {i, n}];       (* radii *)
or = Ordering[rt]; Do[ps[[or[[i]]]] = i, {i, n}];
Do[g[[i]] += x[[i]] * Sign[rt[[i]] - c[[ps[[i]]]]]/n
   ,{i, n}];         (* radii gradient contribution *)
k = 0;
Do[dt[[++k]] = Total[(x[[i]] - x[[j]])^2]/2, {i, 2, n}
   , {j, i - 1}];            (* calculate distances *)
orp = Ordering[dt]; Do[psp[[orp[[i]]]] = i, {i, np}];
k = 0;
Do[k++;           (* distance gradient contribution *)
 ch =2(x[[i]]-x[[j]])*Sign[dt[[k]]-cp[[psp[[k]]]]]/np;
 g[[i]] += ch; g[[j]] -= ch , {i, 2, n}, {j, i - 1}];
x -= alpha * g;            (* gradient descent step *)
\end{verbatim}
\end{footnotesize}

\bibliographystyle{IEEEtran}
\bibliography{cites}
\end{document}